\documentclass[10pt,twocolumn,letterpaper]{article}

\usepackage{iccv}
\usepackage{times}
\usepackage{epsfig}
\usepackage{graphicx}
\usepackage{amsmath}
\usepackage{amssymb}
\usepackage{mathrsfs}
\usepackage{algorithm}
\usepackage{algpseudocode}
\usepackage{graphicx,subcaption}
\usepackage{enumitem}

\usepackage{booktabs}
\usepackage{authblk}

\usepackage[breaklinks=true,bookmarks=false]{hyperref}

\iccvfinalcopy 

\def\iccvPaperID{4365} 

\ificcvfinal\fi

\begin{document}

\title{Seq-SG2SL: Inferring Semantic Layout from Scene Graph \\Through Sequence to Sequence Learning}

\newcommand*\samethanks[1][\value{footnote}]{\footnotemark[#1]}

\author[]{Boren Li\thanks{Corresponding author.}  \qquad Boyu Zhuang \qquad Mingyang Li \qquad Jian Gu\\
{Alibaba AI Labs\\
	\tt\small \{boren.lbr\textsuperscript{*}, po-yu.zby, gujian.gj\}@alibaba-inc.com}\\
{\tt\small mingyangli009@gmail.com}}

\maketitle
\ificcvfinal\thispagestyle{empty}\fi

\begin{abstract}
Generating semantic layout from scene graph is a crucial intermediate task connecting text to image. We present a conceptually simple, flexible and general framework using sequence to sequence (seq-to-seq) learning for this task. The framework, called Seq-SG2SL, derives sequence proxies for the two modality and a Transformer-based seq-to-seq model learns to transduce one into the other. A scene graph is decomposed into a sequence of semantic fragments (SF), one for each relationship. A semantic layout is represented as the consequence from a series of brick-action code segments (BACS), dictating the position and scale of each object bounding box in the layout. Viewing the two building blocks, SF and BACS, as corresponding terms in two different vocabularies, a seq-to-seq model is fittingly used to translate. A new metric, semantic layout evaluation understudy (SLEU), is devised to evaluate the task of semantic layout prediction inspired by BLEU. SLEU defines relationships within a layout as unigrams and looks at the spatial distribution for $n$-grams. Unlike the binary precision of BLEU, SLEU allows for some tolerances spatially through thresholding the Jaccard Index and is consequently more adapted to the task. Experimental results on the challenging Visual Genome dataset show improvement over a non-sequential approach based on graph convolution.
\end{abstract}

\section{Introduction}\label{sec:intro}
Learning the relation from semantic description to its visual incarnation leads to important applications, such as text-to-image synthesis \cite{Zhang_2017_ICCV} and semantic image retrieval \cite{Johnson_2015_CVPR}. It remains a challenging and fundamental problem in computer vision \cite{Johnson_2018_CVPR}. Recent researches have gradually formalized the structured representations of the two modality, scene graph \cite{Johnson_2015_CVPR}\cite{Krishna_2017_IJCV} for semantic description and semantic layout \cite{Hong_2018_CVPR}\cite{Zhao_2019_CVPR} for image. Therefore, our goal in this work solves the underlying task, inferring semantic layout from scene graph, for connecting text to image.
\begin{figure}
	\begin{center}
		\includegraphics[width=8cm]{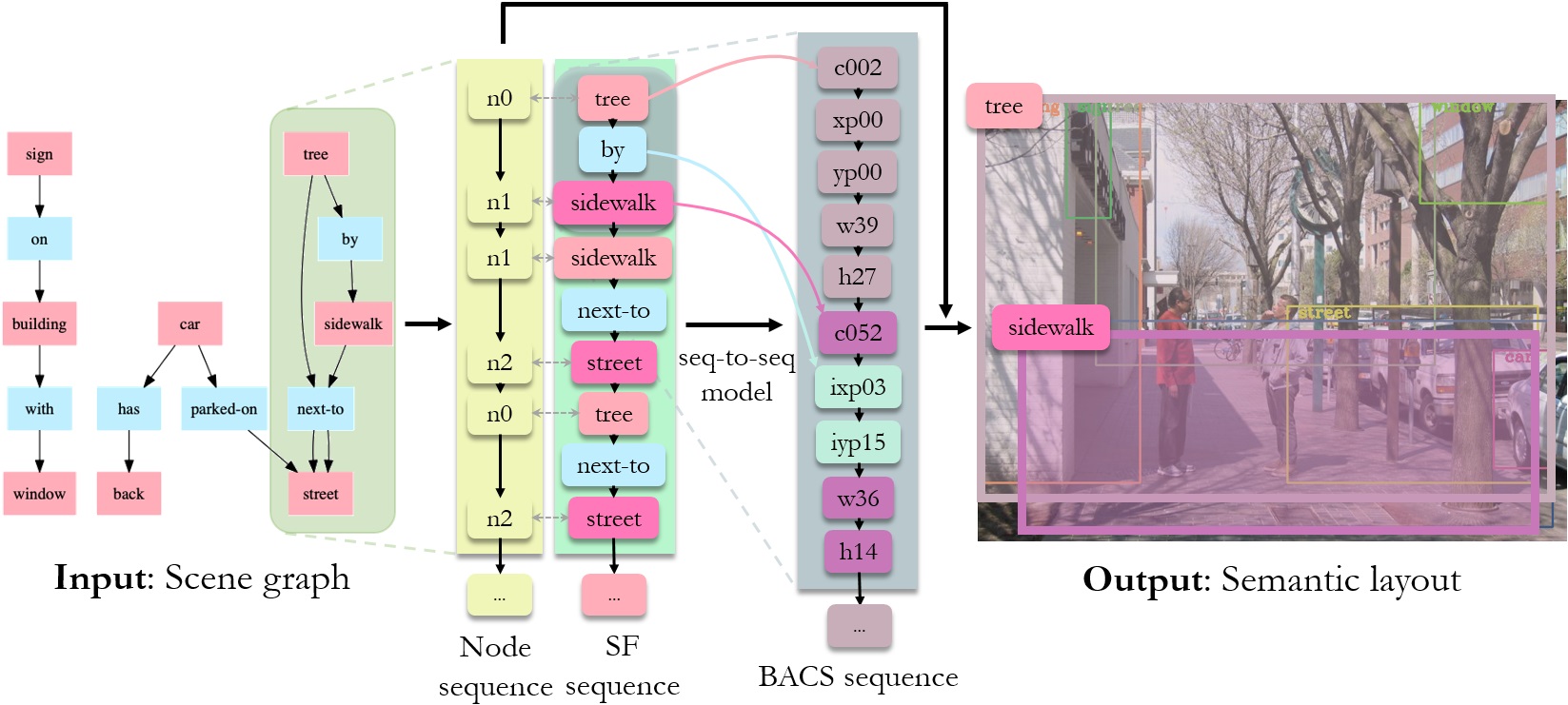}
	\end{center}
	\vspace{-0.4cm}
	\caption{The Seq-SG2SL framework for inferring semantic layout from scene graph.}
	\vspace{-0.5cm}
	\label{fig:Seq-SG2SL}
\end{figure}

Most existing works infer semantic layout from text \cite{Hinz_2019_ICLR}\cite{Hong_2018_CVPR}\cite{Tan_2019_CVPR}. However, leading methods still struggle with complex text inputs depicting multiple objects owing to the unstructured nature of text. Hence, Johnson \etal \cite{Johnson_2018_CVPR} pioneered to infer semantic layout from scene graph as a separate task, isolated from semantic parsing \cite{Schuster_2015_EMNLP}. Scene graph is adopted because it is a powerful structured representation that efficiently conveys scene contents in text \cite{Krishna_2017_IJCV}. As a notable step forward, they observed that semantic layout is largely constrained by objects within relationships. As such, they developed a graph convolution network to embed a scene graph containing only objects within relationships into respective object feature vectors that further dictate semantic layout through an object layout network. However, they inferred all object feature embeddings simultaneously from a scene graph that comprises exponential variability of object and relationship combinations. It is extremely challenging for a model to express such prohibitive diversity.

We view this task from a novel perspective to avoid combinatorial explosion that largely restricted the model expressiveness in the past. Inferring semantic layout from scene graph can be compared to constructing a building from its blueprint. It is unwise to offer corpus of blueprint-to-building correspondences to directly train a learner how to construct a building from its blueprint. Instead, teaching the basic actions to stack the building blocks based on their counterparts in the blueprint is much more feasible. What determines the building blocks? It is the \textit{relationship}.

We propose a conceptually simple, flexible and general framework using sequence to sequence (seq-to-seq) learning to infer semantic layout from scene graph (Figure \ref{fig:Seq-SG2SL}). The framework, called Seq-SG2SL, derives sequence proxies for the two modality and a Transformer-based seq-to-seq model learns to transduce one into the other. A scene graph is decomposed into a sequence of semantic fragments (SF), one for each relationship. A semantic layout is the consequence from a series of brick-action code segments (BACS), dictating the position and scale of each object bounding box in the layout. Viewing the two building blocks, SF and BACS, as corresponding terms in two different vocabularies, a seq-to-seq model is fittingly used to \textit{translate}. Seq-SG2SL is an intuitive framework that learns BACS to drag-and-drop and scale-adjust the two bounding boxes of subject and object in a relationship to the layout supervised by its SF counterpart.

Direct and automated evaluation for semantic layout prediction is another challenging problem unto itself. A new metric, semantic layout evaluation understudy (SLEU), is devised for this purpose inspired by BLEU \cite{Kishore_2002_ACL}. SLEU defines relationships within a layout as unigrams and looks at the spatial distribution for $n$-grams. Unlike the binary precision of BLEU, SLEU allows for some tolerances spatially through thresholding the Jaccard Index and is consequently more adapted to the task. Mean-SLEU over a large corpus is a proper metric for evaluation.

We experiment on the challenging Visual Genome (VG) dataset \cite{Krishna_2017_IJCV} that provides human annotated pairs of scene graph and semantic layout for each image. We first show qualitative results from Seq-SG2SL and rationalize SLEU intuitively. Further quantitative comparison shows the advantages of Seq-SG2SL over a non-sequential approach based on graph convolution \cite{Johnson_2018_CVPR}, especially in the aspect of model expressiveness. We show further that this advantage originates from our sequential formulation, not merely from the Transformer model. Various aspects of Seq-SG2SL are studied extensively from additional ablation experiments.

The key contributions are:
\begin{itemize}[leftmargin=*]
	\vspace{-0.25cm}
	\setlength{\itemsep}{0pt}
	\item Seq-SG2SL is the first framework to infer semantic layout from scene graph through seq-to-seq learning and outperforms the non-sequential state-of-the-art model by a significant margin.
	\item SLEU is the first automatic metric to directly evaluate the performance of semantic layout prediction, allowing results reproducibility.
\end{itemize}

\section{Related Works}\label{sec:related_works}
\noindent\textbf{Scene Graph:} A scene graph is a directed graph representing a scene, where nodes are objects and edges give relationships between objects. Johnson \etal \cite{Johnson_2015_CVPR} first introduced the notion of scene graph as query input for semantic image retrieval. They predicted the most likely semantic layout from a query scene graph as intermediate result for the end retrieval task through a conditional random field (CRF) model. Meanwhile, Schuster \etal \cite{Schuster_2015_EMNLP} complemented their work by introducing an automatic approach to create a scene graph from unstructured natural language scene descriptions. Having these brilliant pioneering attempts using scene graph, Krishna \etal \cite{Krishna_2017_IJCV} constructed the VG dataset that aimed to bridge language and vision using dense image annotations. Scene graph and semantic layout were adopted as intermediate representations for the two modality. With the advent of VG, scene graph further shows its value from successive researches, such as in predicting grounded scene graph for images \cite{Cewu_2016_ECCV}\cite{Xu_2017_CVPR}, in evaluating image captioning \cite{Peter_2016_ECCV}, and in image generation \cite{Johnson_2018_CVPR}.

\noindent\textbf{Semantic Description to Semantic Layout:} Semantic layout was first officially given as spatial distribution of clip arts in abstract scene proposed by Zitnick \etal \cite{Zitnick_2013_CVPR}. This representation initially aimed to study directly for inferring high-level semantics from images. By contrast, Zitnick \etal \cite{Zitnick_2013_ICCV} formulated the opposite problem of predicting an abstract scene from its textual description and proposed a solution using CRF. As clip arts in abstract scene can be easily generalized to object bounding boxes in semantic layout, this concept extends to real images \cite{Tan_2019_CVPR}. 

Predicting a semantic layout from text is usually posed as an intermediate step for complex image generation \cite{Hinz_2019_ICLR}\cite{Hong_2018_CVPR}. A complex image refers to the one containing multiple interactive objects. Unlike the family of methods \cite{Qiao_2019_CVPR}\cite{Zhang_2017_ICCV} that can give stunning results on limited domains, such as fine-grained descriptions of birds or flowers, a semantic layout is usually necessary for complex image generation to dictate multi-object spatial distribution depicted in a text. Johnson \etal \cite{Johnson_2018_CVPR} pioneered to structurize text into scene graph for further complex image generation. This work is the closest to ours. We adopt the same idea but only focus on its subtask of semantic layout prediction from scene graph. The rest of the task, image generation from layout, can be separately solved as \cite{Zhao_2019_CVPR}. In contrast to the closest work that proposed a non-sequential approach based on graph convolution, Seq-SG2SL views the task from a novel perspective and formulates the problem in a seq-to-seq manner.

During quantitative studies, all existing works did not perform direct evaluations on semantic layout prediction. Instead, they applied \textit{indirect metrics}, such as the inception score or the image captioning score derived from the generated image. Though human evaluations are all incorporated for further evaluation, these results are very expensive to be \textit{reproduced}, which logjam the way for fruitful research ideas in this field. The circumstance is exceptionally similar to that before BLEU \cite{Kishore_2002_ACL} was first introduced to the field of machine translation.

\noindent\textbf{Sequence to Sequence Learning:} RNN, LSTM \cite{Hochreiter_1997_NC} and GRU \cite{Chung_2014_CoRR} are firmly established for sequence modelling and transduction problems, such as language modelling and machine translation \cite{Cho_2014_CoRR}\cite{Ilya_2014_NIPS}. Attention mechanisms further become integral parts of compelling sequence modelling and the transduction models, allowing modelling dependencies without regarding to the distance in the input or output sentences \cite{Bahdanau_2014_CoRR}. Vaswani \etal \cite{Vaswani_2017_NIPS} popularized the Transformer architecture that can boost the performance of machine translation by a significant margin. Larger-scale architecture exploration specifically for machine translation was conducted using Transformer to further converge towards the optimal settings  \cite{Britz_2017_CoRR}. Similar conclusion was also drawn by \cite{Klein_2017_CoRR}. Seq-to-seq learning is still evolving rapidly. Beneficial experiences can be borrowed for our purpose.

\section{Seq-SG2SL}\label{sec:Seq-SG2SL}
Seq-SG2SL is conceptually simple: it predicts a series of actions to form a consequent layout from a corresponding sequence of symbolic triplets derived from relationships in a scene graph. Next, we introduce the design of sequence proxies for the two modality that is key in our work.

\subsection{Sequence Proxies}\label{subsec:seq_proxies}
A scene graph encodes objects, attributes and relationships, while its resulting layout is constrained merely by the objects within \textit{relationships}. Therefore, a scene graph is first preprocessed to drop all attributes and independent objects not within any relationship.

A relationship in scene graph is represented by a symbolic triplet: subject, predicate, object. The sequence proxy for a triplet is a tuple, named SF. It is a successive concatenation of the three elements. The preprocessed scene graph can be then decomposed into a sequence of SF, one for each relationship. To fully preserve information in the scene graph, a node sequence, consisting of corresponding object node IDs, is additionally maintained. Note that our Seq-SG2SL framework is flexible to transfer object attributes from a scene graph to a layout through this sequence.

The visual incarnation of an SF in semantic layout contains a pair of object bounding boxes, each for the subject and object. They are referred to as \textit{visual subject} and \textit{visual object}. Its sequence proxy is a BACS. The design requirements are threefold: first, the series of BACS must \textit{uniquely} determine a layout; second, a BACS should correspond to an SF such that the direction of \textit{causality} is clear; third, vocabularies in BACS must be \textit{representable} and \textit{repeatable} such that they can concisely represent any layout.

A semantic layout, space-quantized into $H \times W$ square grids, is called a quantized layout where all BACS are defined. It requires $5$ types of actions to form an object bounding box in the layout: four to specify the location and scale, and the other to set its class index. The bounding box location for the subject is represented in absolute coordinates, whereas that for the object uses the \textit{relative position} to the subject. This relativity aims to encode the \textit{visual predicate} in a relationship. We show by an experiment that this relative position encoding is key for good performance.

\begin{table}
	\begin{center}
		\begin{tabular}{c|c}
			\hline
			Type & Functionality \\
			\hline\hline
			c & set class index of bbox \\
			xp & set xmin of subject bbox \\
			yp & set ymin of subject bbox \\
			ixp & increase xmin of object bbox from subject \\
			ixn & decrease xmin of object bbox from subject \\
			iyp & increase ymin of object bbox from subject \\
			iyn & decrease ymin object bbox from subject \\
			w & set width of bbox \\
			h & set height of bbox \\
			imgar & set aspect ratio index of semantic layout \\
			\hline
		\end{tabular}
	\end{center}
	\vspace{-0.3cm}
	\caption{The BACS types and functionalities. The minimum values for x and y are respectively denoted as xmin and ymin. The bounding box is abbreviated as bbox.}
	\vspace{-0.3cm}
	\label{tab:BACS_types}
\end{table}

The BACS types and functionalities are formalized in Table \ref{tab:BACS_types}. A BACS is composed of $10$ consecutive words and corresponds to a $3$-word SF. Types for the $10$ words are sequentially: c, xp, yp, w, h, c, ixp(n), iyp(n), w, h. The first and last $5$ words in a BACS form the visual subject and object, respectively. A BACS sequence is the direct concatenation of individual BACS with identical relationship order to its corresponding SF sequence. Optionally, imgar is added to the front of a BACS sequence in case the aspect ratio of semantic layout is of interest.

\subsection{Sequence-to-Sequence Model}\label{subsec:seq-to-seq_model}
Given an SF sequence and its corresponding BACS sequence, a seq-to-seq model is fittingly used to translate. We adopt the most recent Transformer model with $6$ stacked self-attention and point-wise, fully connected layers for both the encoder and decoder, exactly identical to the one in \cite{Vaswani_2017_NIPS}. This model is used because of its superior performance in machine translation whose formulation is identical to ours. We show by an experiment that the advantage of model expressiveness by Seq-SG2SL originates from our sequential formulation, not merely from Transformer.

\subsection{Semantic Layout Restoration}\label{subsec:sl_restore}
Having predicted BACS sequence from its input SF sequence, the \textit{alignment} is first verified by checking brick action types for each word sequentially. If aligned, the BACS sequence corresponds to both the input SF and node sequences. The predicted brick actions are then successively executed to form the restored layout. 

Note that with the subsidiary node sequence, it is sufficient to derive which bounding boxes in the restored layout should be merged to one as given in scene graph. If the bounding boxes to be merged have identical predicted class index, the merged bounding box is computed simply as their mean value. Otherwise, it picks the one with median bounding box area. In fact, the predicted class indices for these bounding boxes are rarely distinct. More careful merging strategy is left for future investigation.

\subsection{Implementation Details}\label{subsec:implementaiton_details}
\noindent\textbf{Semantic Layout Encoding:} The maximum side length of a quantized layout is set at $40$. Larger values lead to more BACS vocabularies, making seq-to-seq model harder to generalize. Smaller values, however, result in imprecise bounding box localization and scaling. The chosen value is a trade-off. The aspect ratio of semantic layout is also uniformly quantized. The quantization interval and minimum value are $0.05$ and $0.5$, respectively.

\noindent\textbf{Data Augmentation:} If a scene-graph corresponding layout is valid, its subgraph counterpart, containing a subset of the relationships, is still acceptable. We augment a scene graph to more pairs of sequences in the two modality by applying this property. The concatenation order of relationships in the pairs of sequences is arbitrary, leading to another freedom for augmentation. To balance training data, each scene graph is augmented to at most $50$ correspondences. To limit the maximum length for both sequences, we only preserve at most $9$ relationships in a scene graph.

\noindent\textbf{Training:} We set hyper-parameters exactly as the Transformer work \cite{Vaswani_2017_NIPS}, except the warm-up steps is set at $8000$. We use the Transformer implementation by OpenNMT \cite{Klein_2017_CoRR} for our purpose. We train on a single Tesla P100 GPU for one million iterations ($2$ days) under batch size of $1024$.

\noindent\textbf{Inference:} We use beam search with beam size of $4$ and length penalty $\alpha=0.6$ \cite{Wu_2016_CoRR}. The inference time is about $132$ms on a single Tesla P100 GPU.

\section{SLEU Metric} \label{sec:SLEU}
Our goal is to design an \textit{automatic} metric to \textit{directly quantify} the success of semantic layout prediction from scene graph. The premise behind automatic evaluation is: the closer a prediction is to human-prepared references, the better it is. Thus, the problem becomes: how to design a metric that can measure the \textit{similarity} between a predicted layout against a set of references.

\subsection{Notation}\label{subsec:SLEU_notation}
Let $\mathcal{L} = \left\{\langle\mathit{r}\rangle_k\right\}_{k\in[1, K]}$ denote a layout with $K$ relationships, where $\langle\mathit{r}\rangle_k = \left(\mathbf{s}_k, \mathbf{o}_k\right)$ represents a \textit{visual relationship}. $\mathbf{s}_k$ and $\mathbf{o}_k$ respectively denote visual subject and object, where $\mathbf{s}_k = \left[c_k^{[s]}, \mathbf{b}_k^{[s]}\right]$ and $\mathbf{o}_k = \left[c_k^{[o]}, \mathbf{b}_k^{[o]}\right]$. $c_k$ denotes the class index. $\mathbf{b}_k=\left[x_k, y_k, w_k, h_k\right]$ dictates the bounding box.

Our objective is to evaluate how \textit{close} a predicted layout $\hat{\mathcal{L}}$ is to a collection of reference layouts $\left\{\mathcal{L}_j\right\}_{j\in\left[1, M\right]}$. $\hat{\left(\cdot\right)}$ represents predicted values later on.

\subsection{Metric Design}\label{subsec:metric_design}
SLEU is devised inspired by BLEU \cite{Kishore_2002_ACL} in machine translation. The cornerstone for BLEU is $n$-gram that refers to a contiguous sequence of $n$ items from a text. Here, the item is \text{word}. Evaluating only $n$-grams for machine translation is justified under the Markov assumption that the appearance probability for the current word is merely determined by its previous $\left(n-1\right)$ words, independent from the $n$\textsuperscript{th} one. The $n$-gram concept is generalized for SLEU, where the item is \textit{relationship}, instead of \textit{word}. By analogy to BLEU, evaluating $n$-grams in a semantic layout assumes the placement for a visual relationship depends only on a maximum of $\left(n-1\right)$ other relationships.

SLEU evaluates a semantic layout from two perspectives: intra-relationship adequacy as \textit{unigram accuracy}; and inter-relationship fidelity as \textit{n-gram accuracy}. These accuracies are finally combined to a single-number metric.
\vspace{-0.25cm}
\subsubsection{Unigram Accuracy}\label{subsubsec:unigram_accuracy}
\vspace{-0.15cm}
\begin{algorithm}
	\caption{Unigram accuracy against a single reference}
	\label{alg:unigram_accuracy}
	\textbf{Input:} A predicted layout $\hat{\mathcal{L}}$ and a reference $\mathcal{L}$\newline
	\textbf{Output:} Unigram accuracy $p_1$
	\begin{algorithmic}[1]
		\Function{$f_1$}{$\hat{\mathcal{L}}, \mathcal{L}$}:
		\State set $c=0$\Comment{$c$: count of matches}
		\For{each $\langle\mathit{r}\rangle_k$ in $\mathcal{L}$}\Comment{$K$ iterations}
		\If{$\hat{c}_k^{[s]}$ = $c_k^{[s]}$ \textbf{and} $\hat{c}_k^{[o]}$ = $c_k^{[o]}$}
		\State compute $\tilde{\mathbf{t}}_k=\left[x_k^{[s]} - \hat{x}_k^{[s]},y_k^{[s]} - \hat{y}_k^{[s]}\right]$
		\State $\hat{\mathbf{b}}_k^{[s]}\gets\hat{\mathbf{b}}_k^{[s]}\oplus\tilde{\mathbf{t}}_k$
		\State
		$\hat{\mathbf{b}}_k^{[o]}\gets\hat{\mathbf{b}}_k^{[o]}\oplus\tilde{\mathbf{t}}_k$
		\State compute $J_{k}^{[s]}=IoU\left(\hat{\mathbf{b}}_k^{[s]}, \mathbf{b}_k^{[s]}\right)$
		\State compute $J_{k}^{[o]}=IoU\left(\hat{\mathbf{b}}_k^{[o]}, \mathbf{b}_k^{[o]}\right)$
		\If{$J_{k}^{[o]}\geq T_{IoU}$ \textbf{and} $J_{k}^{[s]}\geq T_{IoU}$}
		\State $c\gets c+1$
		\EndIf
		\EndIf
		\EndFor
		\State \textbf{return} $p_1=c/K$
		\EndFunction
	\end{algorithmic}
\end{algorithm}

The unigram accuracy $p_1$ quantifies matching of individual relationships in a predicted layout to a reference, as shown in Algorithm \ref{alg:unigram_accuracy}. It simply compares the pairs of visual relationships and counts the number of matches. $p_1$ is then the count of the matched-pair divided by the total number of relationships.

To compare a pair of visual relationships, one first needs to compute a shift vector $\tilde{\mathbf{t}}_k$ that aligns the center of $\hat{\mathbf{b}}_k^{[s]}$ to $\mathbf{b}_k^{[s]}$. Then $\hat{\mathbf{b}}_k^{[s]}$ and $\hat{\mathbf{b}}_k^{[o]}$ are center-shifted by this vector, where $\oplus$ denotes this operation. The two Jaccard Indices for the two pairs of bounding boxes are then computed. These Jaccard Indices, each for visual subject and object, give rise to the binary decision of matching through thresholding by $T_{IoU}$ that allows some tolerances spatially.
\vspace{-0.15cm}
\subsubsection{$n$-gram Accuracy}\label{subsubsec:ngram_accuracy}
\vspace{-0.15cm}
\begin{algorithm}
	\caption{$n$-gram accuracy against a single reference}
	\label{alg:ngram_accuracy}
	\textbf{Input:} A predicted layout $\hat{\mathcal{L}}$ and a reference $\mathcal{L}$\newline
	\textbf{Output:} $n$-gram accuracy $p_n$
	\begin{algorithmic}[1]
		\Function{$f_n$}{$\hat{\mathcal{L}}, \mathcal{L}$}:
		\State set $c=0$\Comment{$c$: count of matches}
		\State compute $\mathbb{P}_n(K)$
		\For{each $\mathcal{P}_n$ in $\mathbb{P}_n$}\Comment{$|\mathbb{P}_n|$ iterations}
		\State compute $\tilde{\mathbf{t}}_q$
		\State set $f = 1$\Comment{$f$: flag of match}
		\For{each $\langle\mathit{r}\rangle_i$ in $\mathcal{P}_n$}\Comment{$n$ iterations}
		\State $\hat{\mathbf{b}}_i^{[s]}\gets\hat{\mathbf{b}}_i^{[s]}\oplus\tilde{\mathbf{t}}_q$
		\State compute $J_i^{[s]}=IoU\left(\hat{\mathbf{b}}_i^{[s]}, \mathbf{b}_i^{[s]}\right)$
		\If{$J_i^{[s]}< T_{IoU}$ \textbf{or} $\hat{c}_i^{[s]} \neq  c_i^{[s]}$}
		\State	$f = 0$
		\State \textbf{break}
		\EndIf
		\EndFor
		\If{$f = 1$}
		\State $c\gets c+1$
		\EndIf
		\EndFor
		\State \textbf{return} $p_n = c/|\mathbb{P}_n|$
		\EndFunction
	\end{algorithmic}
\end{algorithm}

The $n$-gram accuracy $p_n$ ($n \in [2, N]$) quantifies similarity of spatial distributions with $n$ visual relationships between the predicted and reference layout, as elaborated in Algorithm \ref{alg:ngram_accuracy}. $\mathbb{P}_n(K)=\{\mathcal{P}_n\}$ denotes a collection that consists of all $n$-relationship subsets of $\mathcal{L}$. $|\cdot|$ represents the cardinality of a set. To compute $p_n$, it first compares each pair of the $n$-relationship subsets and counts the number of matches. $p_n$ is then the matched proportion of these pairs of $n$-relationship subsets.

To measure similarity between a pair of $n$-relationship subsets, we only use their $n$ visual subjects. This is because the relative distribution from a visual object to its subject has been encoded in unigram accuracy. To compare two spatial distributions, each with $n$ visual subjects, a shift vector $\tilde{\mathbf{t}}_q$ is first computed to align the centroids of $\{\hat{\mathbf{b}}_i^{[s]}\}_{i \in [1,n]}$ and $\{\mathbf{b}_i^{[s]}\}_{i \in [1,n]}$. All elements in $\{\hat{\mathbf{b}}_i^{[s]}\}_{i \in [1,n]}$ are then center-shifted by $\tilde{\mathbf{t}}_q$. If all the shifted bounding boxes pass the Jaccard-Index thresholding and class-alignment check, this pair of $n$-relationship subsets is considered match.
\vspace{-0.15cm}
\subsubsection{SLEU Score}\label{subsubsec:SLEU}
\hspace{\parindent}SLEU combines unigram and $n$-gram accuracies as a single-number indicator. For a predicted layout and a reference, the unigram and $n$-gram accuracies are separately computed. Similar to the observation by BLEU, $n$-gram accuracy decays roughly exponentially with $n$. Hence, SLEU adopts the same averaging scheme: a weighted average of logarithm accuracies.

SLEU can also measure similarity between a prediction and \textit{multiple} references. One first needs to compare a prediction with each reference, obtaining a combined accuracy each. Then, the highest value, corresponding to the closest reference to the prediction, is simply designated as SLEU.

SLEU is formally defined as
\begin{equation}\label{eq:SLEU}
{\rm SLEU} = \max_{j\in[1,M]}\left(e^{\sum_{n=1}^{N}w_n{\rm ln}\left(p_n^j\right)}\right),
\end{equation}
where $N=3$ and $w_n=1/N$ as \textit{uniform weights}. $N=3$ is chosen experimentally to make SLEU more distinguishable since larger $n$ leads to negligible small $n$-gram accuracies.

SLEU ranges from $0$ to $1$. Few prediction can attain a score of $1$ unless it is very similar to one of the references. Involving more reasonable reference layouts corresponding to a scene graph leads to a higher value of SLEU. Thus, one must be cautious to \textit{the number of reference} during making comparisons on evaluations. The quality of SLEU can be enhanced by adding the quantity of reference.
\vspace{-0.25cm}
\subsubsection{Mean-SLEU Metric}\label{subsubsec:mean-SLEU}
\hspace{\parindent}One may still wonder the SLEU capability to measure the prediction performance, especially when only one layout is available in the reference corpus. The design of SLEU strictly follows the idea from the well-known BLEU. We just extend the concept from 1D to 2D, which can lead to the binary decision of \textit{visual relationship matching} and allow some tolerances spatially. Therefore, SLEU is largely justified from its similarity to BLEU. 

From the work of BLEU \cite{Kishore_2002_ACL}, using a single-reference test corpus for evaluation is valid if the corpus size is large and the reference translations are not all from the same translator. In our case, the corpus size is maintained at least several thousands and all the samples are crowd-sourced. Hence, the \textit{mean-SLEU} over a large set can be justifiable for evaluation, though its correlation with human judgement is still desirable for future investigation.
\begin{figure*}
	\begin{center}
		\begin{minipage}[b]{0.16\linewidth}
			\includegraphics[width=1.1in]{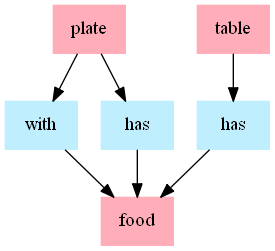}\\
			\includegraphics[width=1.1in,height=1.1in]{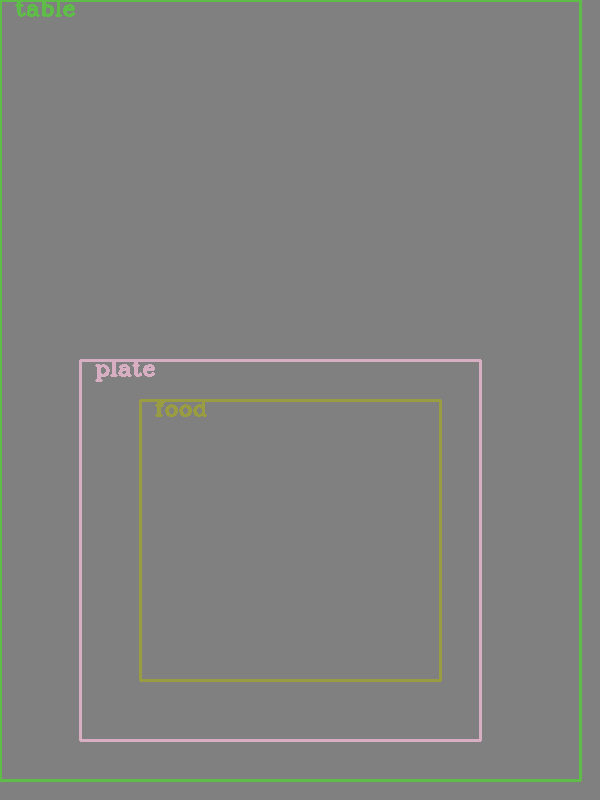}\\
			\includegraphics[width=1.1in,height=1.1in]{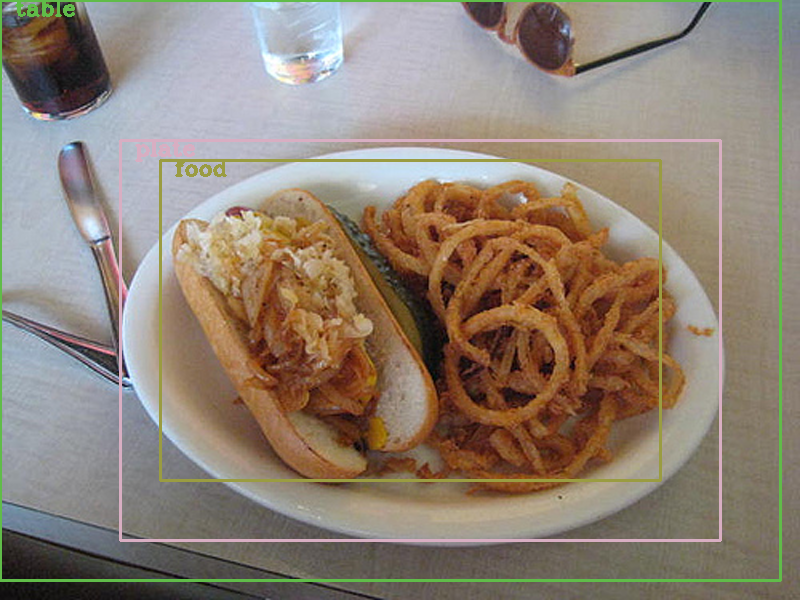}
			\captionsetup{font={small},justification=centering}
			\caption*{(a) $\rm SLEU=0.87$}    
		\end{minipage}
		\begin{minipage}[b]{0.16\linewidth}
			\begin{center}
				\includegraphics[width=1.1in]{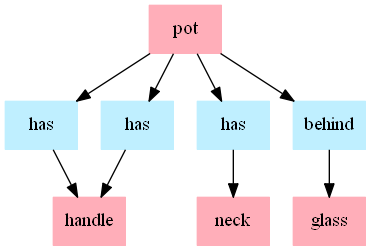}\\
		\end{center}
			\includegraphics[width=1.1in,height=1.1in]{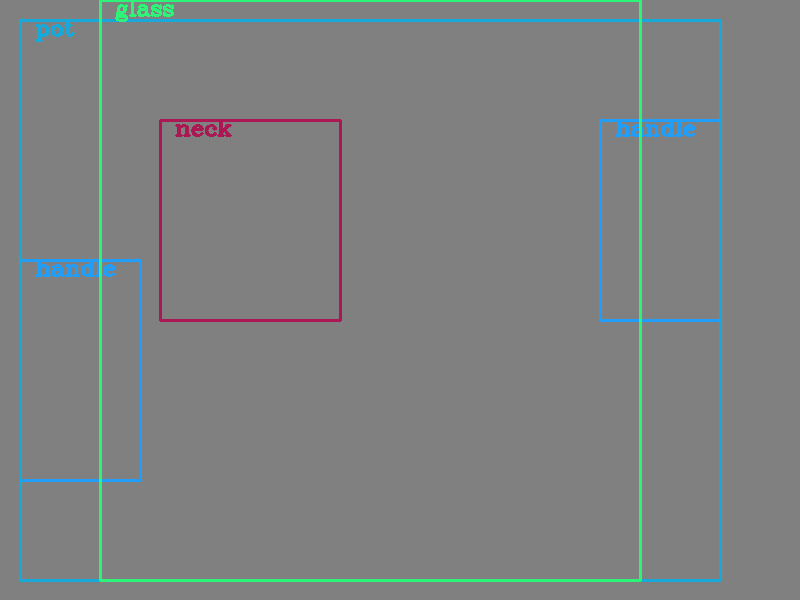}\\
			\includegraphics[width=1.1in,height=1.1in]{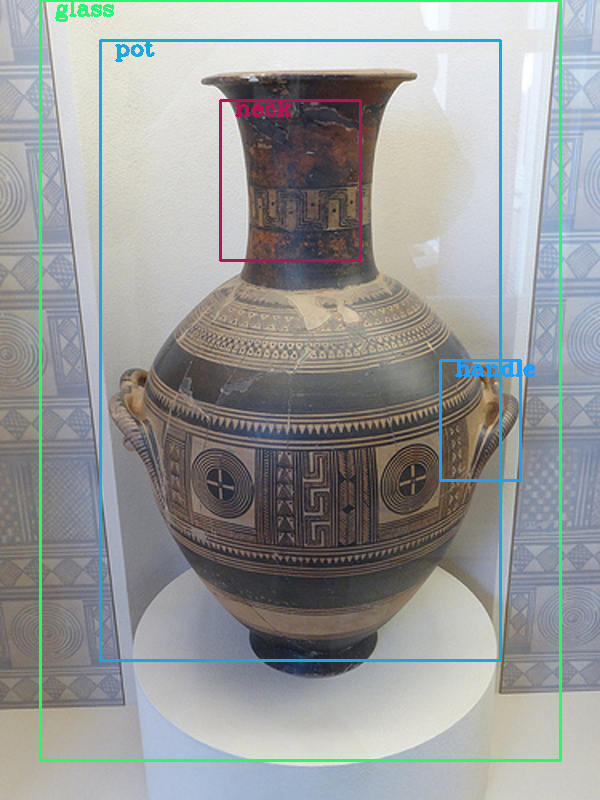}
			\captionsetup{font={small},justification=centering}
			\caption*{(b) $\rm SLEU=0.63$}         
		\end{minipage}
		\begin{minipage}[b]{0.16\linewidth}
			\begin{center}
				\includegraphics[width=1.1in]{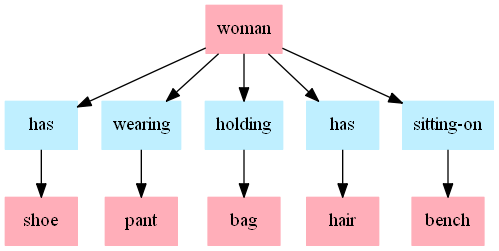}\\
			\end{center}
			\includegraphics[width=1.1in,height=1.1in]{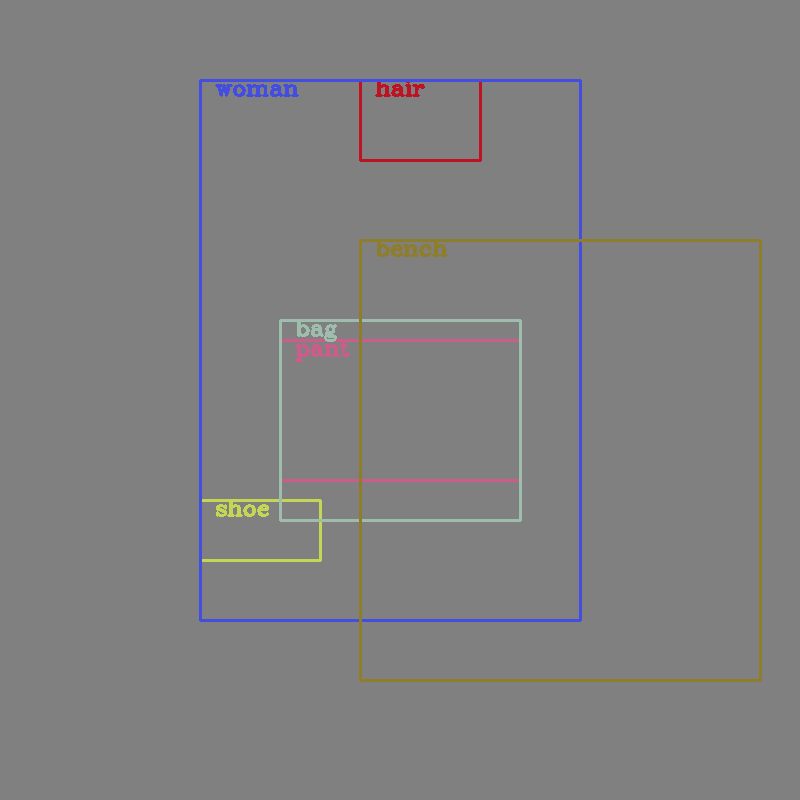}\\
			\includegraphics[width=1.1in,height=1.1in]{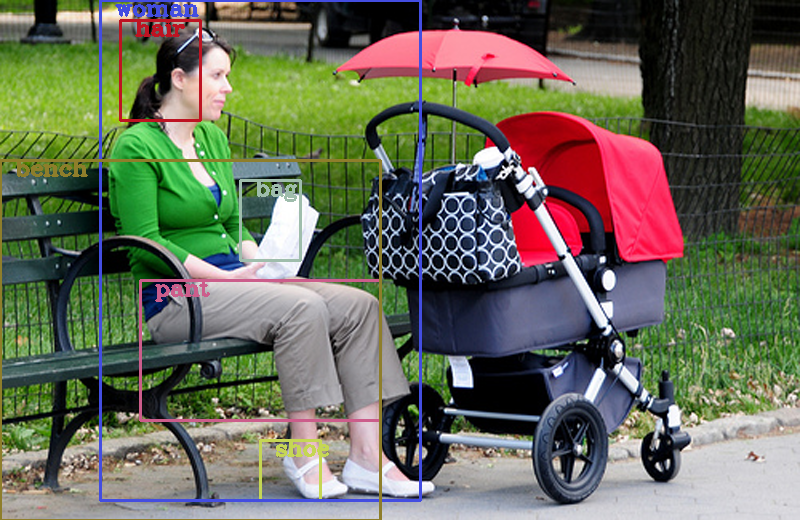}
			\captionsetup{font={small},justification=centering}
			\caption*{(c) $\rm SLEU=0.58$}         
		\end{minipage}
		\begin{minipage}[b]{0.16\linewidth}
			\includegraphics[width=0.95in]{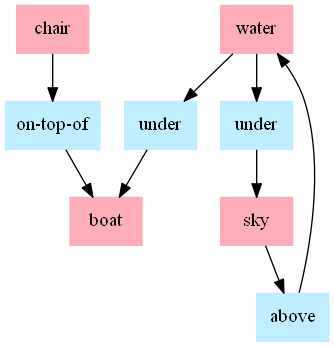}\\
			\includegraphics[width=1.1in,height=1.1in]{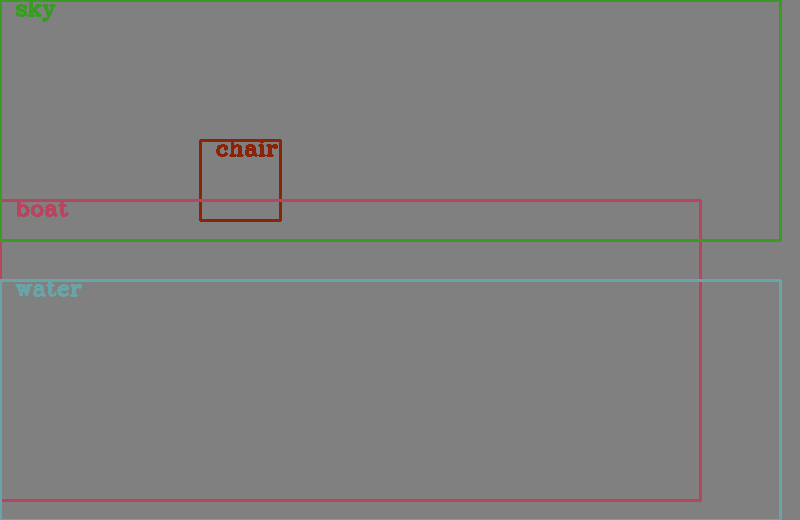}\\
			\includegraphics[width=1.1in,height=1.1in]{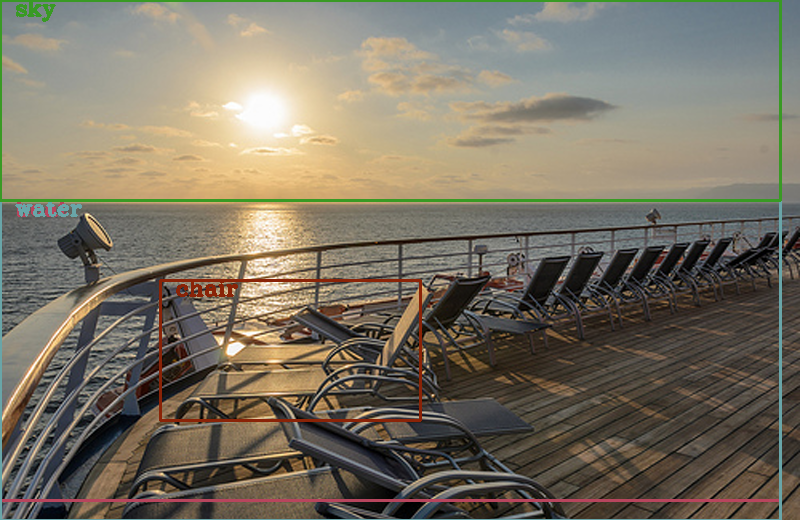}
			\captionsetup{font={small},justification=centering}
			\caption*{(d) $\rm SLEU=0.45$}         
		\end{minipage}
		\begin{minipage}[b]{0.16\linewidth}
			\begin{center}
				\includegraphics[width=1.1in]{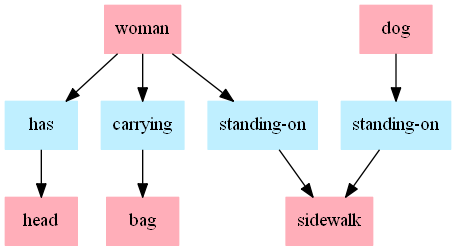}\\
			\end{center}
			\includegraphics[width=1.1in,height=1.1in]{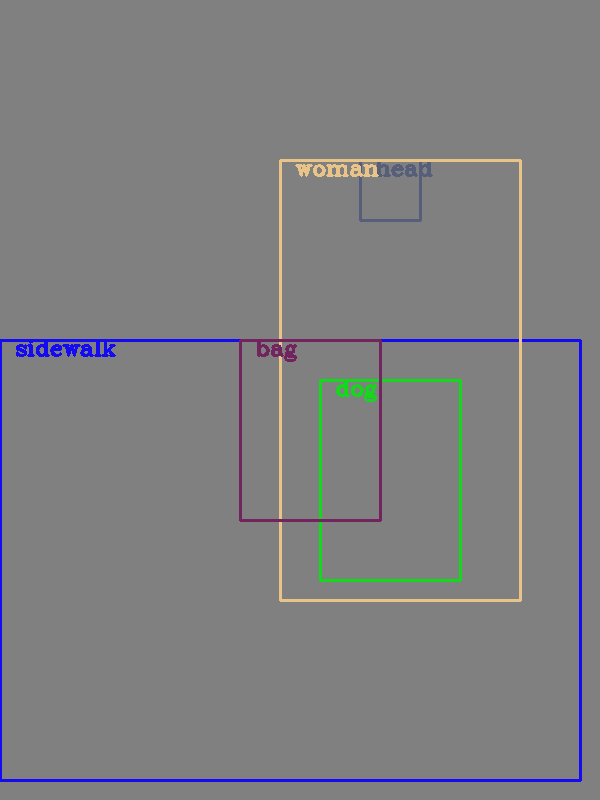}\\
			\includegraphics[width=1.1in,height=1.1in]{"./images/fig_new/0.2/2348791/gt_layout_img".png}
			\captionsetup{font={small},justification=centering}
			\caption*{(e) $\rm SLEU=0.31$}         
		\end{minipage}
		\begin{minipage}[b]{0.16\linewidth}
			\includegraphics[width=1.1in]{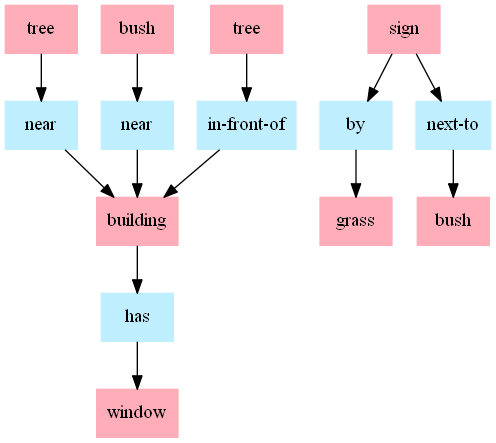}\\
			\includegraphics[width=1.1in,height=1.1in]{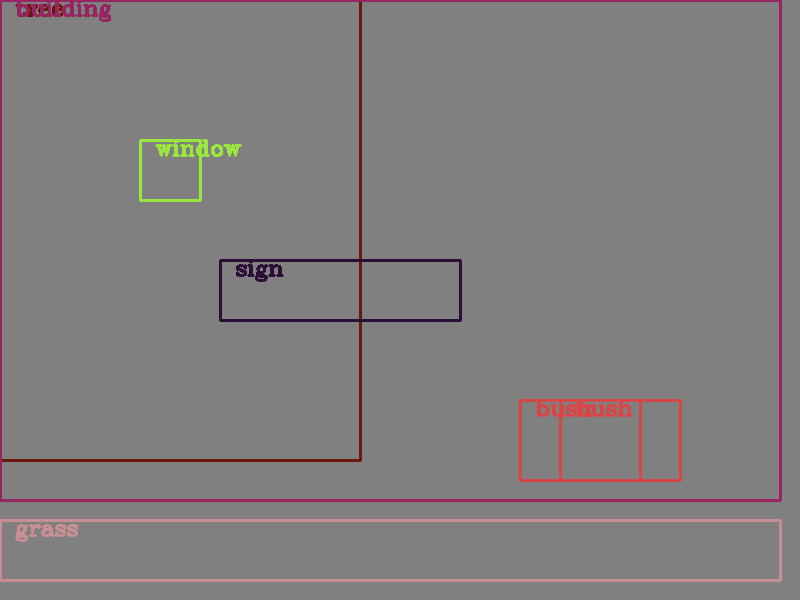}\\
			\includegraphics[width=1.1in,height=1.1in]{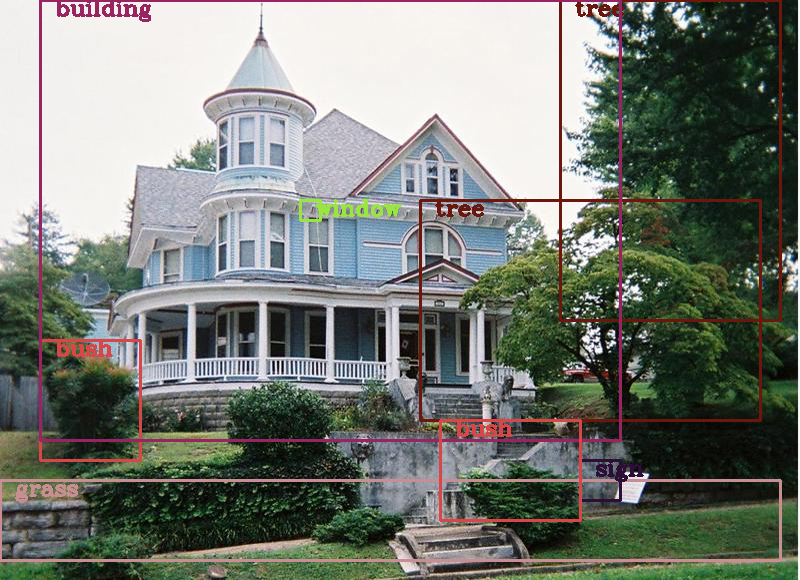}
			\captionsetup{font={small}, justification=centering}
			\caption*{(f) $\rm SLEU=0.00$}         
		\end{minipage}
	\end{center}
	\setlength{\abovecaptionskip}{-1pt}
	\setlength{\belowcaptionskip}{-1pt}
	\caption{Examples of predicted layouts from Seq-SG2SL on the test set of Visual Genome: the first row is the input scene graph; the second row is the predicted layout; the third row is a reference layout overlaid on its corresponding image.}
	\label{fig:quan_iou05}
\end{figure*}
\section{Experiments}\label{sec:exp}
\subsection{Dataset}\label{subsec:dataset}
We experiment on the VG dataset \cite{Krishna_2017_IJCV}. VG comprises $108,077$ images, each contains a scene graph and a corresponding layout. To facilitate later comparison with \cite{Johnson_2018_CVPR}, the dataset is organized exactly following its public implementation. We use the same dataset division: $80\%$ training, $10\%$ validation, and $10\%$ test. We use object and relationship classes occurring at least $2000$ and $500$ times respectively in the training set, leaving $178$ object and $45$ relationship types. We discard tiny bounding boxes with side length shorter than $32$ pixels. We preserve images with between $3$ to $30$ objects and at least one relationship. This leaves us with $62,565$ training, $5,498$ validation, and $5,096$ test images with an average of ten objects and five relationships per image. Different from \cite{Johnson_2018_CVPR}, we limit the maximum number of relationships in a scene graph to $9$ to adapt for our Seq-SG2SL setting. Here, for a scene graph with more relationships, we simply keep the first $9$ while discard the rest.

\subsection{Qualitative Results}\label{subsec:qualitative_results}
This experiment aims to visualize results from Seq-SG2SL and rationalize SLEU intuitively. Figure \ref{fig:quan_iou05} demonstrates these results from test set and their SLEU scores under $T_{IoU}=0.5$. Qualitatively, Seq-SG2SL is capable to predict a semantic layout from a scene graph containing multiple interactive objects.

We analyze by examples from Figure \ref{fig:quan_iou05} to intuitively understand how SLEU can distinguish a good prediction from a bad one. A high SLEU score greater than $0.8$, such as (a), definitely indicates a good prediction since the predicted layout is sufficiently close to one of the references. For those with medium SLEU scores between $0.4$ to $0.8$, as shown in (b)-(d), most predictions are still comparable to a reference. Objects with smaller bounding boxes may appear at apparently irregular positions, such as the \textit{chair} in (d), whereas larger ones are more likely to be reasonable. For cases with low SLEU scores smaller than $0.4$, as shown in (e) and (f), the predictions are basically different from the reference. A low SLEU score either suggests a bad prediction or a reasonable prediction that far deviates from any of the given references. Provided more references, SLEU can be more representable, just as the rationale of BLEU.

\subsection{Quantitative Comparison}\label{subsec:quantitative_comparison}
\begin{figure*}[htp]
	\centering
	\begin{subfigure}{0.95\linewidth}
		\centering
		\includegraphics[width=17cm]{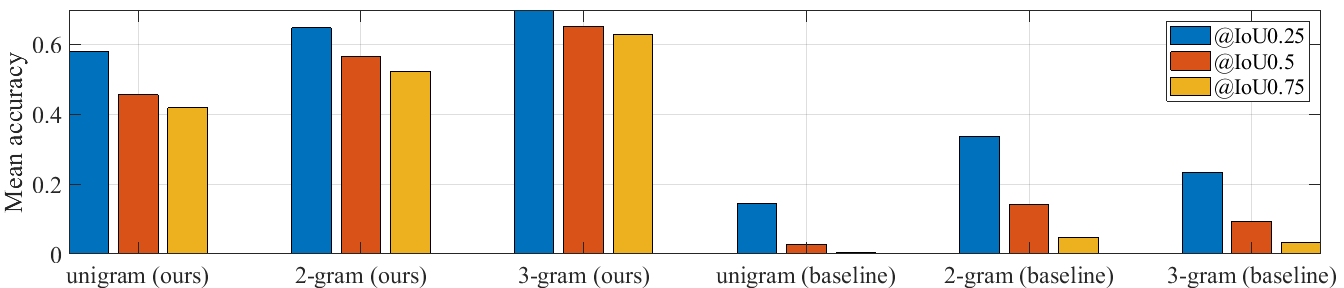}\quad
	\end{subfigure}
	\begin{subfigure}{0.95\linewidth}
		\centering
		\includegraphics[width=17cm]{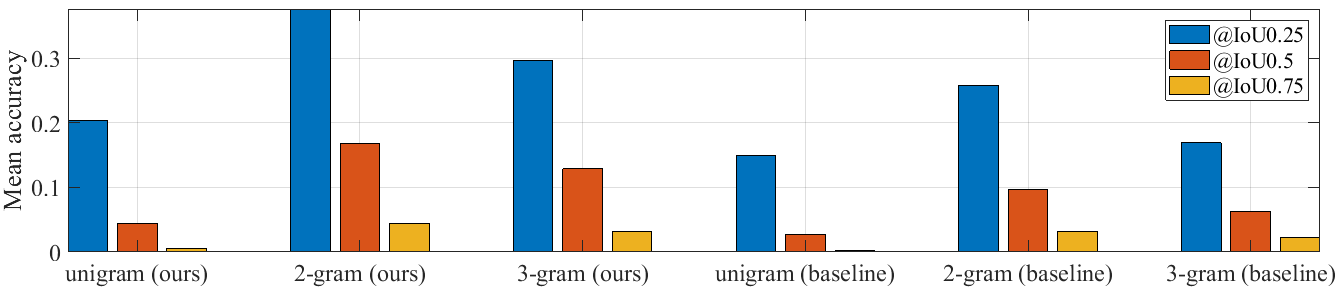}\par\medskip
	\end{subfigure}\textsl{}
	\caption{Comparison of mean n-grams accuracies under different $T_{IoU}$ thresholds between our Seq-SG2SL and the baseline approach on the two sets. The top and bottom figure respectively show results on the training and test set.}
	\label{fig:compare_baseline}
\end{figure*}

This experiment compares Seq-SG2SL with a baseline approach \cite{Johnson_2018_CVPR}. The pretrained model of the baseline is applied to generate layouts for benchmarking. Our Seq-SG2SL model is trained on exactly the same set as the baseline. Mean-SLEU is employed for evaluation on both the training and test set. Each sample in the two sets consists of a single reference layout corresponding to a scene graph.

\begin{table}
	\begin{center}
		\begin{tabular}{c|c|c|c|c}
			\hline
			& IoU$0.0$ & IoU$0.25$ & IoU$0.5$ & IoU$0.75$ \\
			\bottomrule[1pt]
			Baseline-train & $1.0000$ & $0.1209$ & $0.0217$ & $0.0036$ \\
			\textbf{Ours-train} & $\mathbf{1.0000}$ & $\mathbf{0.5012}$ & $\mathbf{0.4174}$ & $\mathbf{0.3942}$\\
			Baseline-test & $\mathbf{1.0000}$ & $0.1163$ & $0.0205$ & $0.0020$\\
			\textbf{Ours-test} & $0.9998$ & $\mathbf{0.1665}$ & $\mathbf{0.0344}$ & $\mathbf{0.0039}$\\
			\hline
		\end{tabular}
	\vspace{-0.2cm}
	\end{center}
	\caption{Comparison of mean-SLEU scores under different $T_{IoU}$ thresholds between the baseline and our Seq-SG2SL approach on the training and test set.}
	\label{tab:comparison_sleu_with_baseline}
\end{table}
Table \ref{tab:comparison_sleu_with_baseline} shows the comparison of mean-SLEU scores under various $T_{IoU}$ thresholds between the two approaches on the training and test set. The larger values, indicating better performances, are highlighted. As demonstrated, Seq-SG2SL outperforms the baseline by a significant margin on the training set. Note that evaluations on the training set are not trivial but offer insight for model expressiveness. By analogy with machine translation, we cannot expect a trained model to produce exactly the same output as a given reference, even on the training set, since the same input inherently results in several different reasonable outputs. But for training, the closer, the better. Therefore, Seq-SG2SL has shown its advantage in model expressiveness over the baseline. This advantage originates from our sequential formulation in avoiding combinatorial explosion. On the test set, Seq-SG2SL still has higher scores than the baseline, suggesting its better generalization capability, except that the score under $T_{IoU}=0.0$ is slightly worse. This score only considers the matching of object classes with no evaluation on the spatial distribution of bounding boxes in the layout. This score from Seq-SG2SL is also very close to $1$, showing its capability in predicting the correct object classes. In rare cases, the predicted BACS sequence cannot find perfect alignment with the input SF sequence, resulting in such negligible gap.

Figure \ref{fig:compare_baseline} demonstrates the mean $n$-grams accuracies for further analyses, where (a) and (b) respectively show comparisons on the training and test set. As shown, Seq-SG2SL outperforms the baseline in each metric. However, the performance gap during training and testing for Seq-SG2SL is phenomenal, which can be explained from two perspectives. First, only a single reference is given for each scene graph during both training and testing. During training, the model is guided to memorize these particular references and turns out to balance performances on all samples. More supervision implies a higher chance the prediction is closer to its reference. That explains why $n$-gram with larger $n$ has better accuracy on the training set. During testing, however, a given single reference may differ significantly from that during training, resulting in an underestimated score. Thus, adding more references for each test sample is one way to mitigate this training-inference gap. Second, this discrepancy also originates from \textit{exposure bias} for seq-to-seq problem. We refer readers to \cite{Ranzato_2015_CoRR} for more detailed explanations. Recent work \cite{Zhang_ACL_2019} may be incorporated in the future to bridge this training-inference gap.

\subsection{Ablation Experiments}\label{subsec:ablation_exp}
\subsubsection{Relative \textit{vs.} Absolute Position Encoding}\label{subsubsec:ablation_relative_position}
\begin{table}
	\begin{center}
		\begin{tabular}{c|c|c|c|c}
			\hline
			& IoU$0.0$ & IoU$0.25$ & IoU$0.5$ & IoU$0.75$ \\
			\bottomrule[1pt]
			Abs-train & $1.0000$ & $0.3760$ & $0.2698$ & $0.2387$ \\
			\textbf{Ours-train} & $\mathbf{1.0000}$ & $\mathbf{0.5012}$ & $\mathbf{0.4174}$ & $\mathbf{0.3942}$\\
			Abs-test & $0.9996$ & $0.1591$ & $0.0334$ & $0.0038$\\
			\textbf{Ours-test} & $\mathbf{0.9998}$ & $\mathbf{0.1665}$ & $\mathbf{0.0344}$ & $\mathbf{0.0039}$\\
			\hline
		\end{tabular}
	\end{center}
	\caption{Comparison of mean-SLEU scores under different $T_{IoU}$ thresholds between encoding methods using absolute and relative position on the training and test set.}
	\label{tab:comparison_sleu_rel}
\end{table}
\hspace{\parindent}Seq-SG2SL applies \textit{relative} position encoding. This relativity aims to represent the \textit{visual predicate} in a relationship. This experiment aims to clarify the significance to use relative position encoding. We compare the results with that from \textit{absolute} position encoding where locations of visual subject and object are represented in absolute coordinates without considering the visual incarnation of predicate. The two models are trained following exactly the same way except the position encoding method. Table \ref{tab:comparison_sleu_rel} shows the results of this comparison. As demonstrated, relative position encoding is much better, suggesting that the training for visual predicate in a relationship is key for good performance.

\subsubsection{Network Architecture}\label{subsubsec:ablation_architecture}
\begin{table}
	\begin{center}
		\begin{tabular}{c|c|c|c|c}
			\hline
			& IoU$0.0$ & IoU$0.25$ & IoU$0.5$ & IoU$0.75$ \\
			\bottomrule[1pt]
			LSTM-train & $1.0000$ & $0.4134$ & $0.3822$ & $0.3719$ \\
			\textbf{Ours-train} & $\mathbf{1.0000}$ & $\mathbf{0.5012}$ & $\mathbf{0.4174}$ & $\mathbf{0.3942}$\\
			LSTM-test & $0.9984$ & $0.1012$ & $0.0222$ & $0.0031$\\
			\textbf{Ours-test} & $\mathbf{0.9998}$ & $\mathbf{0.1665}$ & $\mathbf{0.0344}$ & $\mathbf{0.0039}$\\
			\hline
		\end{tabular}
	\end{center}
	\caption{Comparison of mean-SLEU scores under different $T_{IoU}$ thresholds between our Transformer-based model and an LSTM-based model on the training and test set.}
	\label{tab:comparison_sleu_arc}
\end{table}
\hspace{\parindent}This experiment compares results from our Transformer-based model and a $4$-layer LSTM-based model with attention mechanism as shown in Table \ref{tab:comparison_sleu_arc}. The two models are trained on exactly the same set. As expected, the LSTM-based model performs slightly worse than our Transformer-based model in each metric. However, its performance on the training set is still far better than the non-sequential baseline (See Table \ref{tab:comparison_sleu_with_baseline}). Therefore, the advantage of model expressiveness mainly comes from our Seq-SG2SL framework in avoiding combinatorial explosion, not only originates from the superiority of the Transformer architecture.
\subsubsection{Greedy \textit{vs.} Beam Search}\label{subsubsec:ablation_greedy_beam_search}
\hspace{\parindent}This experiment compares results from greedy search and beam search. Table \ref{tab:comparison_sleu_top4} shows the top-$4$ results from beam search in addition to the greedy decoding baseline. As shown, on the training set, the best mean-SLEU score is in accordance with the top-$1$ choice from beam search, which justifies the use of the top-$1$ prediction. The top-$2$ to top-$4$ scores appear almost the same but distinctively lower than the top-$1$ score. These results may enrich the diversity of semantic layout predictions, though they are more likely to be unreasonable. The score of the top-$1$ prediction from beam search is higher than that from greedy search as expected, showing the advantage to use beam search. On the test set, however, all scores are generally low and resemble each other. The single-reference test corpus may not be sufficient to conclude with these minor discrepancies. 

\begin{table}
	\begin{center}
		\begin{tabular}{c|c|c|c|c|c}
			\hline
			& Greedy & Top-1 & Top-2 & Top-3 & Top-4 \\
			\bottomrule[1pt]
			train & $0.3958$ & $\mathbf{0.4174}$ & $0.3531$ & $0.3396$ & $0.3357$ \\
			test & $0.0335$ & $0.0344$ & $0.0325$ & $0.0343$ & $\mathbf{0.0367}$ \\
			\hline
		\end{tabular}
	\end{center}
	\setlength{\abovecaptionskip}{-1pt}
	\setlength{\belowcaptionskip}{-3pt}
	\caption{Comparison among results from greedy search and beam search using mean-SLEU score under $T_{IoU}=0.5$ from Seq-SG2SL on the training and test set.}
	\label{tab:comparison_sleu_top4}
\end{table}

\section{Conclusions And Future Works}\label{sec:conc_and_future_works}
This paper has presented Seq-SG2SL, a conceptually simple, flexible and general framework for inferring semantic layout from scene graph through seq-to-seq learning. Seq-SG2SL outperforms a non-sequential state-of-the-art approach by a significant margin, especially in the aspect of model expressiveness. This advantage mainly originates from our sequential formulation in avoiding combinatorial explosion. Relative position encoding to represent visual predicate in a relationship is also key for good performance. Beam search during inference can further enhance the quality of prediction compared with greedy search. The top-$1$ result from beam search is the best layout prediction. SLEU, an automatic metric to directly evaluate semantic layout prediction from scene graph, has been devised in a similar fashion to BLEU. The rationale of SLEU and the justification of using mean-SLEU metric over a large set for evaluation have been discussed.

Direct future works are twofold: first, as for SLEU, its quality can be further enhanced by adding more references. Human evaluations on SLEU may also be of interest. More evaluation metrics on similar tasks, such as text-to-layout synthesis, may be derived by adopting our $n$-gram analogy regarding \textit{relationship} as unigram; second, as for Seq-SG2SL, alternative BACS designs may be investigated, such as altering the parameterization of bounding box similar to CenterNet \cite{Duan_2019_CoRR}. Additive layout generation may also be investigated due to the sequential nature of our framework.

{\small
\bibliographystyle{ieee_fullname}
\bibliography{egbib}
}

\end{document}